\pdfoutput=1

\documentclass[11pt]{article}

\usepackage[final]{acl}
\usepackage{booktabs} 
\usepackage{multirow}
\usepackage{times}
\usepackage{latexsym}
\usepackage{inconsolata}
\usepackage[table]{xcolor}
\usepackage{amsmath}
\usepackage{colortbl}
\usepackage{multirow}
\usepackage{siunitx}
\usepackage{listings}
\usepackage{xcolor}
\usepackage{url}

\definecolor{codegreen}{rgb}{0,0.6,0}
\definecolor{codegray}{rgb}{0.5,0.5,0.5}
\definecolor{codepurple}{rgb}{0.58,0,0.82}
\definecolor{backcolour}{rgb}{0.95,0.95,0.92}

\lstdefinestyle{pythonstyle}{
    backgroundcolor=\color{backcolour},   
    commentstyle=\color{codegreen},
    keywordstyle=\color{magenta},
    numberstyle=\tiny\color{codegray},
    stringstyle=\color{codepurple},
    basicstyle=\ttfamily\footnotesize,
    breakatwhitespace=false,         
    breaklines=true,                 
    captionpos=b,                    
    keepspaces=true,                 
    numbers=left,                    
    numbersep=5pt,                  
    showspaces=false,                
    showstringspaces=false,
    showtabs=false,                  
    tabsize=2
}

\usepackage[T1]{fontenc}
\usepackage[utf8]{inputenc}

\usepackage{microtype}

\usepackage{inconsolata}

\usepackage{graphicx}

\usepackage{xspace}

\usepackage{cleveref}

\usepackage{todonotes}
\usepackage{xcolor}
\usepackage{fancyvrb}
\usepackage{fvextra}

\usepackage{amssymb}

\usepackage{CJKutf8}

\usepackage[breakable]{tcolorbox}

\usepackage{fancyvrb}
\DefineVerbatimEnvironment{Verbatim}{Verbatim}{fontfamily=cmr}
\definecolor{LightTeal}{RGB}{168, 216, 216}  %
\definecolor{LightGold}{RGB}{255, 210, 157}  %
\usepackage{xcolor}

\usepackage{acro}

\usepackage{caption}

\newcommand{\mybenchmark}{{\sf PricingLogic}\xspace}

\definecolor{calmblue}{RGB}{100, 149, 237} %

\title{PricingLogic: Evaluating LLMs Reasoning on Complex Tourism Pricing Tasks}

\author{
Yunuo Liu$^{1}$, Dawei Zhu$^{2}$, Zena Al-Khalili$^{2}$, Dai Cheng$^{3}$, Yanjun Chen$^{3,4}$\\ \textbf{Dietrich Klakow $^{2}$, Wei Zhang$^{3}$, Xiaoyu Shen$^{3}$}\thanks{Corresponding Author}\\
$^{1}$ Hunan University 
$^{2}$ Saarland University, Saarland Informatics Campus \\
$^{3}$ Ningbo Key Laboratory of Spatial Intelligence and Digital, Derivative\\
Institute of Digital Twin, EIT\\
$^{4}$ Department of Computing, The Hong Kong Polytechnic University \\
\texttt{s2302w0374@hnu.edu.cn},
\texttt{xyshen@eitech.edu.cn}
}

\begin{document}
\maketitle
\begin{abstract}

We present \mybenchmark, the first benchmark that probes whether Large Language Models (LLMs) can reliably automate tourism-related prices when multiple, overlapping fare rules apply. Travel agencies are eager to offload this error-prone task onto AI systems; however, deploying LLMs without verified reliability could result in significant financial losses and erode customer trust. \mybenchmark comprises 300 natural-language questions based on booking requests derived from 42 real-world pricing policies, spanning two levels of difficulty: (i) basic customer-type pricing and (ii) bundled-tour calculations involving interacting discounts. Evaluations of a line of LLMs reveal a steep performance drop on the harder tier, exposing systematic failures in rule interpretation and arithmetic reasoning. These results highlight that, despite their general capabilities, today’s LLMs remain unreliable in revenue-critical applications without further safeguards or domain adaptation. Our code and dataset are available at \url{https://github.com/EIT-NLP/PricingLogic}.

\end{abstract}

\section{Introduction}

Recent advances in Large Language Models (LLMs) have demonstrated remarkable capabilities across diverse domains, such as code generation~\cite{chen2021evaluating, chen2022program, hui2024qwen25coder}, mathematical problem-solving~\cite{hendrycks2021measuring,ahn2024large}, and general-purpose human instruction following~\cite{zhou2023instructionfollowing, chen2024benchmarking, chiang2024chatbotarenaopenplatform}. However, real-world deployment remains challenging, as practical applications require domain-specific knowledge, navigation of conflicting rules, and high reliability in contexts where error tolerance is minimal. These requirements are not fully captured by existing benchmarks~\cite{zhou_rulearena_2024}.

\begin{figure*}[!t]
    \centering
    \includegraphics[height=0.45\textwidth]{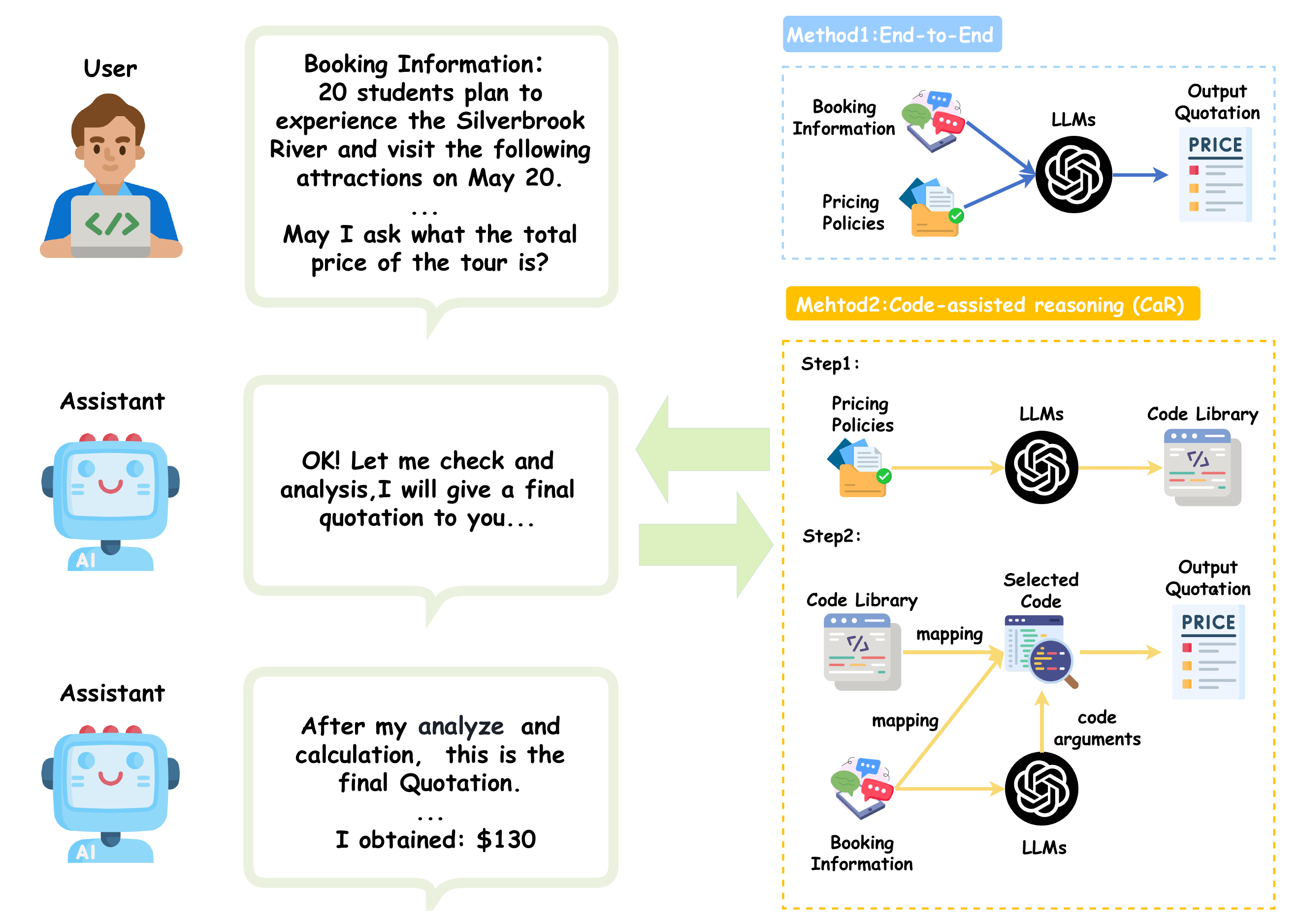}
    \caption{Automatic quotation use case (left) and its two LLM-based realizations (right).}
    \label{fig:task_illustration}
\end{figure*}

In this paper, we focus on a specific yet representative real-world task: automating pricing calculations for tourism bookings, in collaboration with travel agencies interested in using LLM-based systems to process questions expressed in \textit{natural language} (\Cref{fig:task_illustration}, left). These questions often involve multiple destinations, varied fare types, and dynamic pricing policies, making manual processing labor-intensive and error-prone. For LLMs, the task is also nontrivial, as it requires reasoning over complex constraints~\cite{jiang2023followbench}.

To systematically evaluate LLMs on this problem, we introduce \mybenchmark, a benchmark specifically designed to evaluate the capabilities of LLMs in handling realistic booking scenarios. We collected 42 real-world pricing policy documents and 300 questions. These questions cover two main tasks: basic customer-type pricing and more advanced bundled-tour calculations, presented in increasing levels of difficulty. Notably, in addition to standard prompting approaches, we also investigate code-assisted reasoning, which has been shown to enhance LLM performance on computational and logical tasks~\cite[i.a.]{chen2022program, pal, lyu2023faithful}. In our approach, LLMs are first prompted to translate pricing policies into executable Python code. For each incoming question written in natural language, the model extracts relevant information and converts it into input arguments for the generated code, which then calculates the price. We find that this method significantly improves accuracy; nevertheless, challenges remain for complex questions (see \Cref{sec:experiments}).Our main contributions are as follows: (1) We introduce \textbf{PriceLogic}, the first comprehensive benchmark for evaluating LLMs on real-world tourism pricing, comprising 300 questions derived from 42 actual pricing policy documents from travel agencies; (2) We perform a thorough evaluation on a range of open-weight and proprietary LLMs on PriceLogic, and show that it indeed poses a significant challenge to LLMs, where state-of-the-art models answer barely more than half of the questions correctly in our most difficult subset; (3) We demonstrate that code-assisted reasoning significantly improves model performance on our benchmark, which requires complex reasoning and computation, suggesting a promising direction for future work to tackle these types of tasks.

\section{PricingLogic Construction}

In this section, we introduce \mybenchmark, a benchmark for evaluating the reasoning abilities of LLMs in tourism pricing scenarios. \mybenchmark comprises 300 questions divided into three difficulty levels (simple, medium, and challenging) with increasing demands on reasoning and computational capabilities. Simple questions involve single customer types with basic pricing rules, such as \textit{``What is the total price for 3 students visiting Eiffel Tower?''}. Medium questions incorporate multiple variables, including groups of more than 10 visitors, mixed demographics, accommodation status, and two to three service combinations. Challenging questions present complex scenarios with large groups (25–55 visitors), diverse demographic compositions, region-specific pricing, multiple attractions, and overlapping discount conditions. Benchmark statistics are provided in Table~\ref{tab:dataset_overview}.

\begin{table}[htbp]
\centering
\begin{tabular}{lc}
\toprule
Category & Count \\
\midrule
\multicolumn{2}{l}{\textbf{Data Collection}} \\
Individual attractions & 33 \\
Bundled attractions & 9 \\
\midrule
\multicolumn{2}{l}{\textbf{Difficulty Distribution (per task)}} \\
Simple questions & 60 \\
Medium questions & 50 \\
Challenging questions & 40 \\
\bottomrule
\end{tabular}
\caption{\mybenchmark data statistics.}
\label{tab:dataset_overview}
\end{table}

\subsection{Collection and Organization of Tourism Products and Discount Policies}

We collected \mybenchmark through partnerships with travel agencies serving 7 scenic areas with 33 distinct activities. We documented pricing policies for nine customer types (regular visitors, contracted groups, seniors, students, etc.), capturing specific pricing structures, discount thresholds, and special conditions (accommodation benefits, combination incentives). This process revealed the complex conditional rules where prices vary based on customer categories and qualifications. We classified policies by location, activity type, client type, and conditions to generate realistic benchmarking questions.

\subsection{Dataset and Task Setups}

\mybenchmark includes two tasks of increasing complexity, described as follows.

\paragraph{Task 1: Standard price policies.}

Task 1 evaluates LLMs' ability to compute the total cost of tourism bookings using 33 pricing documents. Bundled packages are excluded from this task. We created 150 test examples with clearly defined parameters: visitor classification (regular, contract, etc.), demographic thresholds (at least 80\% students or at least 70\% seniors), group size requirements (10 or more for group rates),  and regional pricing variations, etc.

\paragraph{Task 2: Bundled price policies.}

Task 2 builds upon Task 1 by introducing bundled-tour discounts, which increase the problem's complexity.  Multiple feasible pricing options (regular and preferential) may apply.  This setup mirrors real-world tourism dynamics, where specific combinations of attractions receive preferential rates (lower total price than booking each attraction separately).

\section{Methods}
\label{sec:method}
We consider two approaches for applying LLMs to tasks in \mybenchmark. The first is prompting, which serves as the most straightforward baseline. We include it to assess how well recent LLMs can solve tasks in \mybenchmark without relying on external tools. The second approach allows LLMs to use external tools, in this case, a Python interpreter, to assist with price computation. Both methods are described as follows.

\paragraph{End-to-end prompting (E2E).} Our E2E approach processes pricing in a single inference pass. We standardized the structure and terminology of pricing policy documents, defined and annotated customer types for each project, and explicitly specified the corresponding prices. The prompt guides LLMs through two stages: (1) identifying project details, visitor counts, and special conditions, and (2) calculating prices based on applicable policies, including accommodation exemptions and combination requirements. The full prompt is provided in \Cref{fig:e2e_prompt}.

\paragraph{Code-assisted reasoning (CaR).} Our approach employs a two-stage procedure for automated price calculation. In the first stage, LLMs generate dedicated calculator functions for each pricing policy file, encapsulating conditional rules for customer categories, discounts, and exemptions. In the second stage, natural language orders are parsed to identify the requested items, retrieve the corresponding calculator functions, extract parameter values (e.g., visitor counts, ticket types, or special conditions), and execute the functions to obtain the final price.

\section{Experiments}
\label{sec:experiments}
\subsection{Experimental Setups} 

\paragraph{Models.} We benchmark a line of recent LLMs including both proprietary ones and open-weight ones, including GPT-4o~\cite{gpt4o}, DeepSeek-V3/R1~\cite{deepseekv3, deepseekr1}, and Claude Sonnet 4~\cite{claude4systemcard},and Qwen2.5-7B/32B/Max~\cite{qwen25}.\footnote{Model versions: \texttt{GPT-4o-0129}, \texttt{DeepSeek-V3-0324}, \texttt{Qwen-Max-1015}, \texttt{claude-sonnet-4-20250514-thinking}.}

\begin{table*}[t]
    \centering
    \renewcommand\tabcolsep{21pt}  %
    \renewcommand{\arraystretch}{0.6}  %

    \begin{tabular}{lllll}
        \toprule
        \multirow{2}{*}{\scalebox{0.9}{\textbf{Inference settings}}} & 
        \multirow{2}{*}{\scalebox{0.9}{\textbf{Model}}} & \multicolumn{3}{c}{\scalebox{0.9}{\textbf{Question difficulties}}} \\
        \cmidrule(lr){3-5}
         & & \scalebox{0.9}{Simple}  
         & \scalebox{0.9}{Medium}  
         & \scalebox{0.9}{Challenging}  \\
        \midrule
        \multirow{6}{*}{\scriptsize E2E}
         & \scriptsize Qwen2.5-7B & \scriptsize 63.33 & \scriptsize 12.00 & \scriptsize 0.00 \\
         & \scriptsize Qwen2.5-32B & \scriptsize 86.67 & \scriptsize 40.00 & \scriptsize 50.00 \\
         & \scriptsize Qwen2.5-Max & \scriptsize 90.00 & \scriptsize 54.00 & \scriptsize 32.50 \\
         & \scriptsize DeepSeek-V3 & \scriptsize 83.33 & \scriptsize 70.00 & \scriptsize 40.00 \\  
         & \scriptsize DeepSeek-R1 & \scriptsize 78.33 & \scriptsize72.00 & \scriptsize 45.00 \\
         & \scriptsize GPT-4o & \scriptsize 81.67 & \scriptsize 58.00 & \scriptsize\textbf{52.50} \\  
         & \scriptsize Claude Sonnet 4 & \scriptsize \textbf{ 91.67} & \scriptsize \textbf{76.00} & \scriptsize\textbf{52.50} \\          
        \midrule
        \multirow{6}{*}{\scriptsize CaR}
         & \scriptsize Qwen2.5-7B & \scriptsize 66.66$^{\color{red}\text{\tiny 3.3$\uparrow$}}$ & \scriptsize 28.00$^{\color{red}\text{\tiny 16.0$\uparrow$}}$ & \scriptsize 5.00$^{\color{red}\text{\tiny 5.0$\uparrow$}}$ \\
         & \scriptsize Qwen2.5-32B & \scriptsize 93.33$^{\color{red}\text{\tiny 6.7$\uparrow$}}$ & \scriptsize 68.00$^{\color{red}\text{\tiny 28.0$\uparrow$}}$ & \scriptsize 35.00$^{\color{teal}\text{\tiny 15.0$\downarrow$}}$ \\
         & \scriptsize Qwen2.5-Max & \scriptsize 92.00$^{\color{red}\text{\tiny 2.0$\uparrow$}}$ & \scriptsize78.00$^{\color{red}\text{\tiny 24.0$\uparrow$}}$ & \scriptsize 55.00$^{\color{red}\text{\tiny 22.5$\uparrow$}}$ \\
         & \scriptsize DeepSeek-V3 & \scriptsize 90.00$^{\color{red}\text{\tiny 6.7$\uparrow$}}$ & \scriptsize 70.00$^{\color{gray}\text{\tiny 0.0$\uparrow$}}$ & \scriptsize 50.00$^{\color{red}\text{\tiny 10.0$\uparrow$}}$ \\ 
         & \scriptsize DeepSeek-R1 & \scriptsize 93.33$^{\color{red}\text{\tiny 15.0$\uparrow$}}$ & \scriptsize 74.00$^{\color{red}\text{\tiny 2.0$\uparrow$}}$ & \scriptsize 57.50$^{\color{red}\text{\tiny 12.5$\uparrow$}}$ \\
         & \scriptsize GPT-4o & \scriptsize\textbf{96.67}$^{\color{red}\text{\tiny 15.0$\uparrow$}}$ & \scriptsize 72.00$^{\color{red}\text{\tiny 14.0$\uparrow$}}$ & \scriptsize 55.00$^{\color{red}\text{\tiny 2.5$\uparrow$}}$ \\
         & \scriptsize Claude Sonnet 4 & \scriptsize\textbf{96.67}$^{\color{red}\text{\tiny 5.0$\uparrow$}}$ & \scriptsize \textbf{80.00}$^{\color{red}\text{\tiny 4.0$\uparrow$}}$ & \scriptsize \textbf{60.00}$^{\color{red}\text{\tiny 7.5$\uparrow$}}$ \\
             
        \midrule
        \multirow{6}{*}{\scriptsize CaR-Oracle}
         & \scriptsize Qwen2.5-7B & \scriptsize 15.00$^{\color{teal}\text{\tiny 51.7$\downarrow$}}$ & \scriptsize 6.00$^{\color{teal}\text{\tiny 22.0$\downarrow$}}$ & \scriptsize 0.00$^{\color{teal}\text{\tiny 5.0$\downarrow$}}$ \\
         & \scriptsize Qwen2.5-32B & \scriptsize 96.67$^{\color{red}\text{\tiny 3.3$\uparrow$}}$ & \scriptsize 92.00$^{\color{red}\text{\tiny 24.0$\uparrow$}}$ & \scriptsize 30.00$^{\color{teal}\text{\tiny 5.0$\downarrow$}}$ \\
         & \scriptsize Qwen2.5-Max & \scriptsize\textbf{100.00}$^{\color{red}\text{\tiny 8.0$\uparrow$}}$ & \scriptsize 82.50$^{\color{red}\text{\tiny 4.5$\uparrow$}}$ & \scriptsize\textbf{55.00}$^{\color{red}\text{\tiny 0.0$\uparrow$}}$ \\
         & \scriptsize DeepSeek-V3 & \scriptsize\textbf{100.00}$^{\color{red}\text{\tiny 10.0$\uparrow$}}$ & \scriptsize 85.00$^{\color{red}\text{\tiny 15.0$\uparrow$}}$ & \scriptsize 52.50$^{\color{red}\text{\tiny 2.5$\uparrow$}}$ \\
         & \scriptsize DeepSeek-R1 & \scriptsize\textbf{100.00}$^{\color{red}\text{\tiny 6.7$\uparrow$}}$ & \scriptsize\textbf{92.50}$^{\color{red}\text{\tiny 18.5$\uparrow$}}$ & \scriptsize\textbf{55.00}$^{\color{red}\text{\tiny 2.5$\uparrow$}}$ \\
         & \scriptsize GPT-4o & \scriptsize 96.67$^{\color{gray}\text{\tiny 0.0$\uparrow$}}$ & \scriptsize 85.00$^{\color{red}\text{\tiny 13.0$\uparrow$}}$ & \scriptsize 50.00$^{\color{red}\text{\tiny 0.0$\uparrow$}}$ \\
         & \scriptsize Claude Sonnet 4 & \scriptsize\textbf{100.0}$^{\color{red}\text{\tiny 3.33$\uparrow$}}$ & \scriptsize 88.00$^{\color{red}\text{\tiny 8.0$\uparrow$}}$ & \scriptsize \textbf{62.50}$^{\color{red}\text{\tiny 2.5$\uparrow$}}$ \\         
        \bottomrule
    \end{tabular}
    \caption{Task 1 results across inference settings. Arrows show performance changes between consecutive settings: CaR vs. E2E (second row vs. first row) and CaR-Oracle vs. CaR (third row vs. second row). Values represent percentage point differences. CaR-Oracle uses human-verified code to isolate parameter extraction errors from code generation issues.}
    \label{tab:task1_performance}
\end{table*}

\paragraph{Inference settings.} As outlined in ~\Cref{sec:method}, we evaluate LLMs using both E2E and CaR approaches. CaR has two potential failure modes: (1) generating incorrect calculation code and (2) invoking code with incorrect parameters. To isolate error sources, we introduce \textbf{CaR-Oracle}, where we manually implement Python code for all pricing policies. In this control condition, LLMs only need to pass correct parameters to human-verified code, enabling precise diagnosis of model limitations by controlling for code quality. We set the temperature to 0.0 across all models for deterministic outputs.

\paragraph{Metrics.} We use exact match to compare the model predictions with the correct answer, and report the accuracy.

\begin{table*}[t]
    \centering
    \renewcommand\tabcolsep{21pt}  %
    \renewcommand{\arraystretch}{0.6}  %

    \begin{tabular}{lllll}
        \toprule
        \multirow{2}{*}{\scalebox{0.9}{\textbf{Inference settings}}} & 
        \multirow{2}{*}{\scalebox{0.9}{\textbf{Model}}} & \multicolumn{3}{c}{\scalebox{0.9}{\textbf{Question difficulties}}} \\
        \cmidrule(lr){3-5}
         & & \scalebox{0.9}{Simple}  
         & \scalebox{0.9}{Medium}  
         & \scalebox{0.9}{Challenging}  \\
        \midrule
        \multirow{6}{*}{\scriptsize E2E}
         & \scriptsize Qwen2.5-7B & \scriptsize 68.33 & \scriptsize 28.00 & \scriptsize 0.00 \\
         & \scriptsize Qwen2.5-32B & \scriptsize 76.67 & \scriptsize 46.00 & \scriptsize 27.50 \\
         & \scriptsize Qwen2.5-Max & \scriptsize 85.00 & \scriptsize 48.00 & \scriptsize 22.50 \\
         & \scriptsize DeepSeek-V3 & \scriptsize 83.33 & \scriptsize 45.00 & \scriptsize 27.50 \\  
         & \scriptsize DeepSeek-R1 & \scriptsize 88.33 & \scriptsize 40.00 & \scriptsize 30.00 \\
         & \scriptsize GPT-4o & \scriptsize 90.00 & \scriptsize 54.00 & \scriptsize 27.50 \\  
         & \scriptsize Claude Sonnet 4 & \scriptsize \textbf {91.67} & \scriptsize \textbf{60.00} & \scriptsize\textbf{35.00} \\          
        \midrule
        \multirow{6}{*}{\scriptsize CaR}
         & \scriptsize Qwen2.5-7B & \scriptsize 61.67$^{\color{teal}\text{\tiny 6.7$\downarrow$}}$ & \scriptsize 22.00$^{\color{teal}\text{\tiny 6.0$\downarrow$}}$ & \scriptsize 12.50$^{\color{red}\text{\tiny 12.5$\uparrow$}}$ \\
         & \scriptsize Qwen2.5-32B & \scriptsize 76.67$^{\color{gray}\text{\tiny 0.0$\uparrow$}}$ & \scriptsize 42.00$^{\color{teal}\text{\tiny 4.0$\downarrow$}}$ & \scriptsize 22.50$^{\color{teal}\text{\tiny 5.0$\downarrow$}}$ \\
         & \scriptsize Qwen2.5-Max & \scriptsize 93.33$^{\color{red}\text{\tiny 8.3$\uparrow$}}$ & \scriptsize 78.00$^{\color{red}\text{\tiny 30.0$\uparrow$}}$ & \scriptsize 35.00$^{\color{red}\text{\tiny 12.5$\uparrow$}}$ \\
         & \scriptsize DeepSeek-V3 & \scriptsize 91.67$^{\color{red}\text{\tiny 8.3$\uparrow$}}$ & \scriptsize 68.00$^{\color{red}\text{\tiny 23.0$\uparrow$}}$ & \scriptsize 30.00$^{\color{red}\text{\tiny 2.5$\uparrow$}}$ \\ 
         & \scriptsize DeepSeek-R1 & \scriptsize 93.33$^{\color{red}\text{\tiny 5.0$\uparrow$}}$ & \scriptsize 70.00$^{\color{red}\text{\tiny 30.0$\uparrow$}}$ & \scriptsize 35.00$^{\color{red}\text{\tiny 5.0$\uparrow$}}$ \\
         & \scriptsize GPT-4o & \scriptsize\textbf{95.00}$^{\color{red}\text{\tiny 5.0$\uparrow$}}$ & \scriptsize 76.00$^{\color{red}\text{\tiny 22.0$\uparrow$}}$ & \scriptsize 37.50$^{\color{red}\text{\tiny 10.0$\uparrow$}}$ \\
         & \scriptsize Claude Sonnet 4 & \scriptsize 93.33$^{\color{red}\text{\tiny 1.67$\uparrow$}}$ & \scriptsize \textbf{80.00}$^{\color{red}\text{\tiny 20.0$\uparrow$}}$ & \scriptsize \textbf{42.50}$^{\color{red}\text{\tiny 7.5$\uparrow$}}$ \\         
        \bottomrule
    \end{tabular}
    \caption{Task 2 results across inference settings. Arrows show CaR performance changes compared to E2E baseline (second row vs. first row). Values represent percentage point differences. CaR demonstrates substantial improvements across most models, particularly on medium difficulty questions with 10-20\% gains.}
    \label{tab:task2_performance}
\end{table*}

\subsection{Results on Task 1}
\Cref{tab:task1_performance} presents model performance on Task 1. For simple questions, all LLMs except Qwen2.5-7B correctly answer more than 76\% of the time under direct prompting (E2E). However, performance declines as question complexity increases. Upon inspection, we find that models frequently misidentify customer categories and/or overlook pricing conditions. For challenging questions, all LLMs barely exceed 50\% accuracy.

The CaR approach improves accuracy over E2E in the vast majority of cases. On simple questions, performance gaps between models narrow, and all except Qwen2.5-7B exceed 90\% accuracy. On challenging questions, CaR also provides substantial gains in most cases, but absolute performance remains below 60\% for all models, leaving considerable room for improvement. Overall, CaR demonstrates the effectiveness of this two-stage inference framework with external tools. A notable outlier is Qwen2.5-32B, where CaR underperforms E2E by 15\%. Further analysis reveals that this model often fails to transform booking information into complete and correct function arguments.

Results from CaR-Oracle shed light on CaR’s failure modes. On simple questions, most models improve further, with three LLMs achieving 100\% accuracy using oracle code. This indicates that strong LLMs can generate accurate code for solving simple tasks but may still miss edge cases in implementation. For medium-difficulty tasks, generated code often contains substantial flaws, though strong LLMs can still map questions to correct code arguments. For challenging tasks, oracle code offers little improvement: even with human-written code, models fail to supply correct arguments, indicating that deep task comprehension remains the main bottleneck.

An interesting case arises with Qwen2.5-7B, which shows substantial degradation with oracle code. We find that the model tends to produce simple code to solve tasks, whereas human-written code is more complex in order to cover corner cases. As a result, the model struggles to interpret these more elaborate implementations and fails to map the correct arguments to them.

\subsection{Results on Task 2}

\Cref{tab:task2_performance} presents the model performance on Task 2. With E2E prompting, even the strongest model, Claude Sonnet 4, achieves only 35.0\% accuracy on challenging questions, demonstrating the difficulty introduced by having to reason about bundled-discount interactions. The CaR approach shows substantial improvements for most models across difficulty levels. Particularly notable are the gains for Qwen2.5-Max, DeepSeek-V3, and DeepSeek-R1 on medium-difficulty questions, with improvements of 30\%, 23\%, and 30\%, respectively. 

The CaR approach's success demonstrates that separating policy interpretation from parameter extraction improves handling of complex pricing logic. Error analysis reveals that models struggle with two specific challenges in Task 2: (1) identifying when bundled discounts should override other customer-type pricing, and (2) calculating the optimal combination when multiple valid bundle options exist.

\subsection{0-shot(E2E) vs 3-shot Evaluation}
We further investigate the impact of prompting strategies by comparing 0-shot prompting which corresponding to our E2E method—with 3-shot performance on representative models, examining whether in-context learning can mitigate the observed performance gaps on complex tourism pricing tasks.

For the 3-shot evaluation, we provided 3 examples with correct answers as in-context demonstrations and evaluated performance on 15 representative test examples: 10 simple questions and 5 medium/challenging questions. Table ~\ref{tab:zero_few_shot} presents the comparison between 0-shot and 3-shot performance across different difficulty levels. The results reveal several important findings: 3-shot prompting shows only marginal improvements on simple questions (1.6\% for GPT-4o, 6.7\% for DeepSeek-R1) with no gains on complex scenarios. This indicates that the observed performance limitations reflect fundamental reasoning challenges rather than insufficient demonstrations, supporting our findings about LLMs’ difficulties with complex pricing tasks.

\begin{table}[h]
\centering
\resizebox{\linewidth}{!}{%
\begin{tabular}{lccc}
\toprule
\textbf{Model/Method} & \textbf{Simple} & \textbf{Medium} & \textbf{Challenging} \\
\midrule
GPT-4o 0-shot & 81.7 & 58.0 & 52.5 \\
GPT-4o 3-shot & 83.3 & 58.0 & 52.5 \\
DeepSeek-R1 0-shot & 78.3 & 72.0 & 45.0 \\
DeepSeek-R1 3-shot & 85.0 & 72.0 & 45.0 \\
\bottomrule
\end{tabular}
} %
\caption{0-shot vs 3-shot Performance Comparison.}
\label{tab:zero_few_shot}
\end{table}

\section{Related work}
\paragraph{LLMs in real-world scenarios.} 
Recent research has focused on evaluating LLMs in real-world applications.
\citet{miserendino2025swe} benchmarked LLMs for freelance software engineering, and \citet{huang2024planning} assessed their tool utilization in real-world scenarios.
Closely related to our work is RuleArena \citep{zhou_rulearena_2024} that tests LLMs' rules-following in real-world domains. Unlike RuleArena's linear difficulty scaling (e.g., increasing bag count) and minimal rule conflicts. Our benchmark evaluates LLMs' ability to select optimal pricing among multiple overlapping conditions across diverse demographics, accommodation status, and service combinations, requiring sophisticated comprehension to identify the most favorable option among competing discount rules.

\paragraph{Code-assisted reasoning.}

Assisting LLMs with code improved their reasoning on computation-intensive tasks, 
\citep{lyu2023faithful}, through generating programmatic steps executed by external interpreters. Methods either employ pure code \citep{chen2022program, pal}, code-language interleaving \citep{lyu2023faithful}, code with algebraic expressions, \citep{imani2023mathprompter}, or code with specialized libraries \citep{das2024mathsensei}. While previous work targeted controlled mathematical problems, our approach extends this paradigm to real-world tourism pricing, exceeding textbook problem complexity, through a two-stage pipeline addressing practical constraints such as diverse customer groups and overlapping discount rules. 

\section{Conclusion}

We introduced \mybenchmark, a benchmark evaluating LLMs on complex tourism pricing tasks. Our experiments show code-assisted reasoning generally outperforms end-to-end approaches, yet even advanced models struggle with challenging pricing scenarios involving multiple overlapping rules. These findings highlight the gap between theoretical reasoning capabilities and practical deployment needs in revenue-critical applications, emphasizing the importance of rigorous evaluation before implementing AI in financial contexts.

\section*{Limitations}

We focused only on E2E prompting and CaR methods for evaluating LLMs on pricing tasks. While fine-tuning LLMs specifically for tourism pricing could potentially improve performance, it would require substantial training data, more computational resources, and retraining whenever pricing policies change—making it impractical in dynamic business environments. Our methods offer some flexibility while still providing meaningful performance benchmarks.

\section*{Acknowledgments}
We thank EIT and IDT High Performance Computing Center for providing computational resources for this project.
This work was supported by the 2035 Key Research and Development Program of Ningbo City under Grant No. 2025Z034.
\bibliography{custom}

\appendix

\section{Prompts}
\label{sec:appendix:prompt}

\begin{figure}[h]
    \centering
    \includegraphics[height=0.6\textwidth]{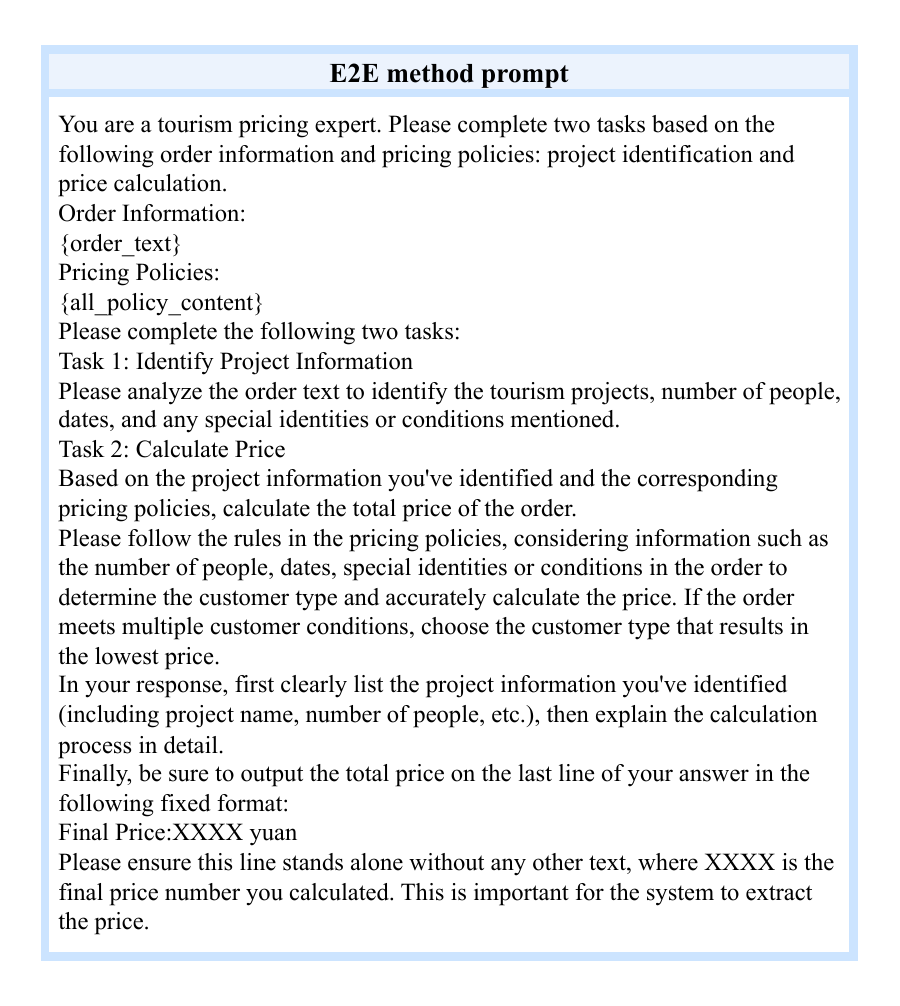}
    \caption{E2E method prompt.}
    \label{fig:e2e_prompt}
\end{figure}

\begin{figure}[h]
    \centering
    \includegraphics[height=0.6\textwidth]{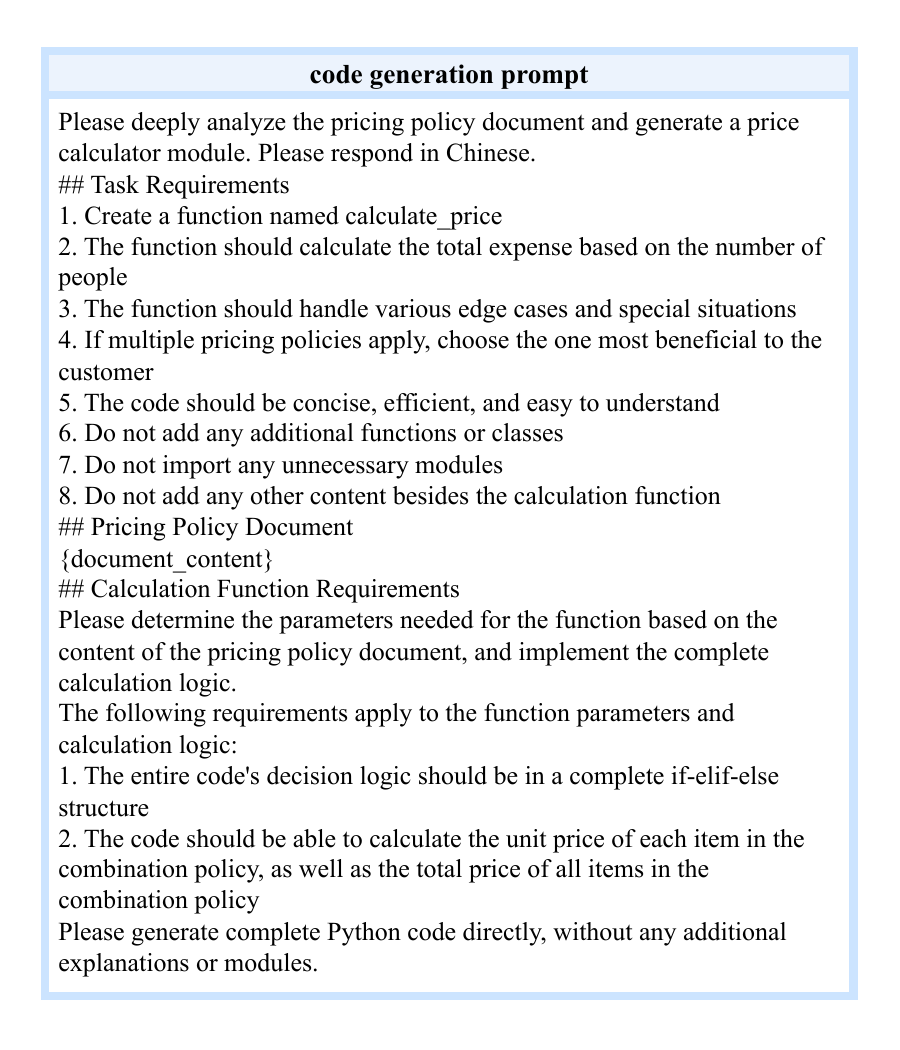}
    \caption{CaR method step1 prompt.}
    \label{fig:car_prompt1}
\end{figure}

\begin{figure}[ht]
    \centering
    \includegraphics[height=0.3\textwidth]{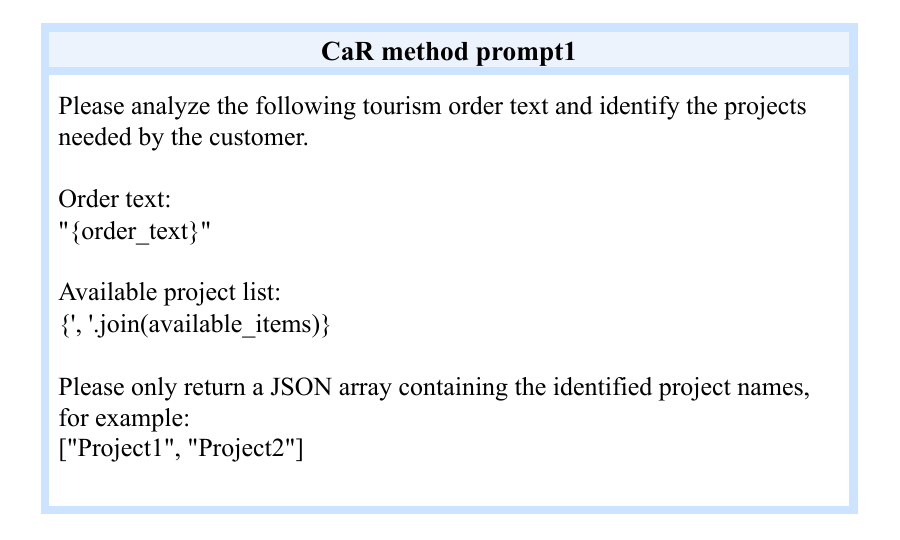}
    \caption{CaR method step2 prompt for booking information analysis.}
    \label{fig:car_prompt2}
\end{figure}

\begin{figure}[ht]
    \centering
    \includegraphics[height=0.7\textwidth]{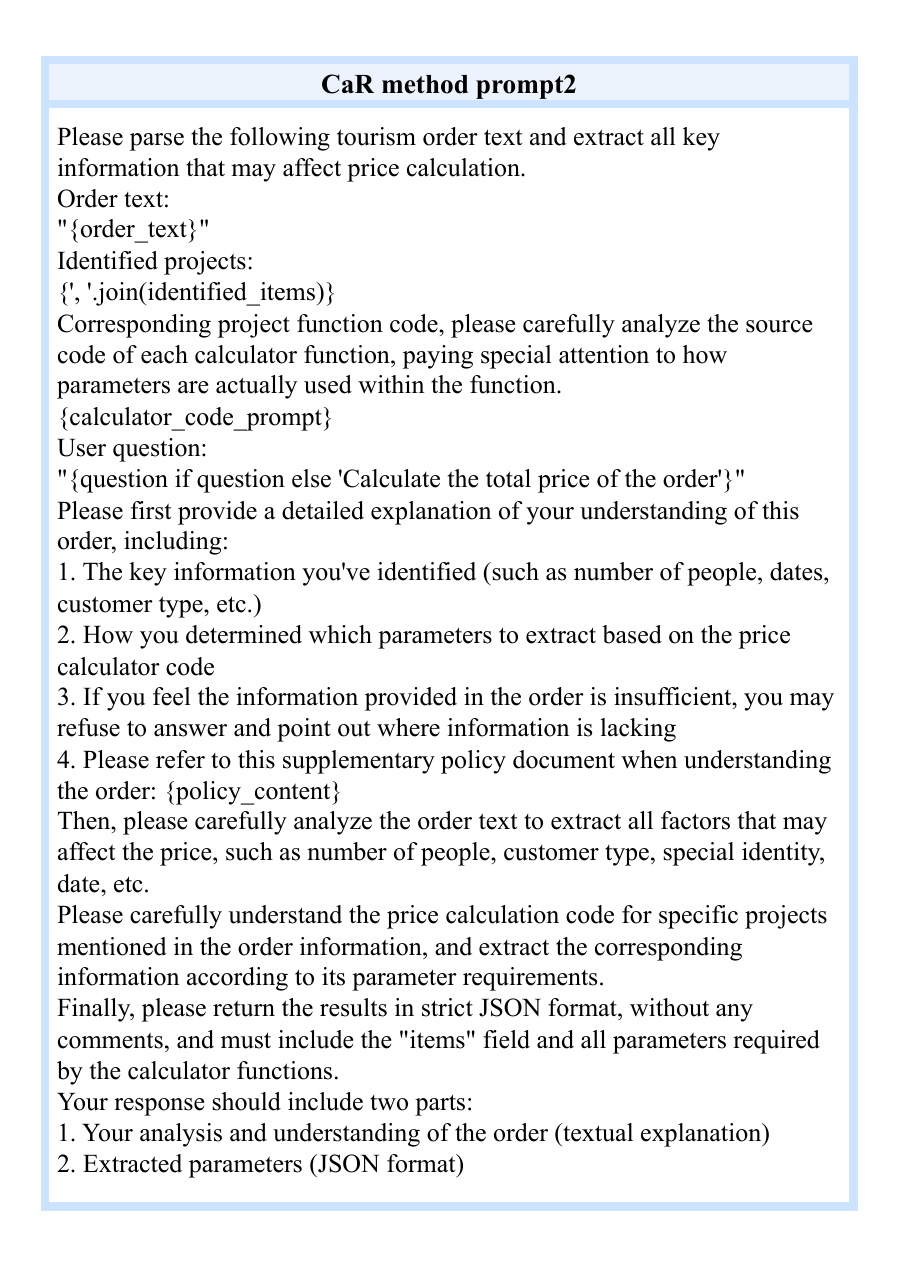}
    \caption{CaR method step2 prompt for code arguments analysis.}
    \label{fig:car_prompt3}
\end{figure}

The E2E prompt is shown in \Cref{fig:e2e_prompt}. The E2E prompt employs a structured approach that guides LLMs through project identification and price calculation, incorporating domain-specific constraints such as hotel partnership benefits and mandatory ticket bundles. By instructing models to select the most favorable pricing option and providing structured output format, this single-pass design tests whether LLMs can handle the full complexity of tourism pricing without task decomposition.
The CaR prompt is shown in \Cref{fig:car_prompt1,fig:car_prompt2,fig:car_prompt3}.
The CaR approach employs a three-stage pipeline: first generating specialized calculator functions from pricing policies (\Cref{fig:car_prompt1}), then identifying relevant projects from questions (\Cref{fig:car_prompt2}), and finally extracting precise parameters guided by the generated code (\Cref{fig:car_prompt3}). The code generation stage enforces customer-favorable pricing logic with priority-ordered conditional structures, while the subsequent stages leverage this structured representation to systematically decompose complex pricing scenarios into manageable computational steps.
The Benchmark example information is shown in \Cref{tab:benchmark_examples}.

\begin{table}[h]
\centering
\begin{tabular}{p{0.95\columnwidth}}
\toprule
\textbf{Benchmark Examples} \\  
\midrule
\textbf{Example of a Simple Question:} \\
3 non-contract customers plan to experience the Harrenstadt Bay tour route. What is the total price for the Harrenstadt Bay tour route? \\
\midrule
\textbf{Example of a Medium Question:} \\
12 tourists (non-contract customers from Essex) are visiting Brighton Cave and St. Elvi Ancient Village. They plan to experience the Brighton Cave entrance ticket and St. Elvi Ancient Village entrance ticket. What is the total price for the Brighton Cave entrance ticket and St. Elvi Ancient Village entrance ticket? \\
\midrule
\textbf{Example of a Challenging Question:} \\
25 tourists (contract customers, including 12 students and 6 seniors, staying at a designated hotel in Clayton Castle) are visiting Brighton Cave, St. Elvi Ancient Village, and Montfiel Monastery. They plan to experience the Brighton Cave entrance ticket, Brighton Cave boat ride, Brighton Cave magic carpet ascent, St. Elvi Ancient Village entrance ticket, and Montfiel Monastery entrance ticket. What is the total price for the Brighton Cave entrance ticket, Brighton Cave boat ride, Brighton Cave magic carpet ascent, St. Elvi Ancient Village entrance ticket, and Montfiel Monastery entrance ticket? \\
\bottomrule
\end{tabular}
\caption{Benchmark Examples Across Difficulty Levels}
\label{tab:benchmark_examples}
\end{table}

\section{Annotation Process}

The \mybenchmark dataset was manually collected and annotated by the authors over a four-day period. We first established clear definitions for various customer types and documented their corresponding pricing structures for each tourism attraction. Table~\ref{tab:customer type price} illustrates an example of the pricing policy for a specific attraction, showing how prices vary across different customer categories. Table~\ref{tab:customer_types} provides detailed definitions of these customer categories, explaining the qualifications and conditions for each pricing tier.

\section{Computing Infra}
Experiments in this work were conducted on a mixed infrastructure setup, with some models run locally and others accessed via API endpoints.
For open-source models, experiments were conducted with different GPU configurations. Qwen2.5-7B was run on a single Nvidia A800 GPU card (80GB), while Qwen2.5-32B required 4 A800 GPUs for inference. The server was equipped with Intel(R) Xeon(R) Platinum 8378A CPU @ 3.00GHz processors. Batch processing was implemented to optimize throughput across all experimental runs.
For larger proprietary models (Qwen2.5-Max, DeepSeek-V3, DeepSeek-R1, GPT-4o, and Claude Sonnet 4), we utilized their respective API endpoints. The API calls were managed through a queuing system to handle rate limits and ensure reliable data collection. All API requests were executed with temperature set to 0.0 to ensure deterministic outputs.

\section{Code Analysis}
\begin{figure}[h]
    \centering
    \includegraphics[height=0.4\textwidth]{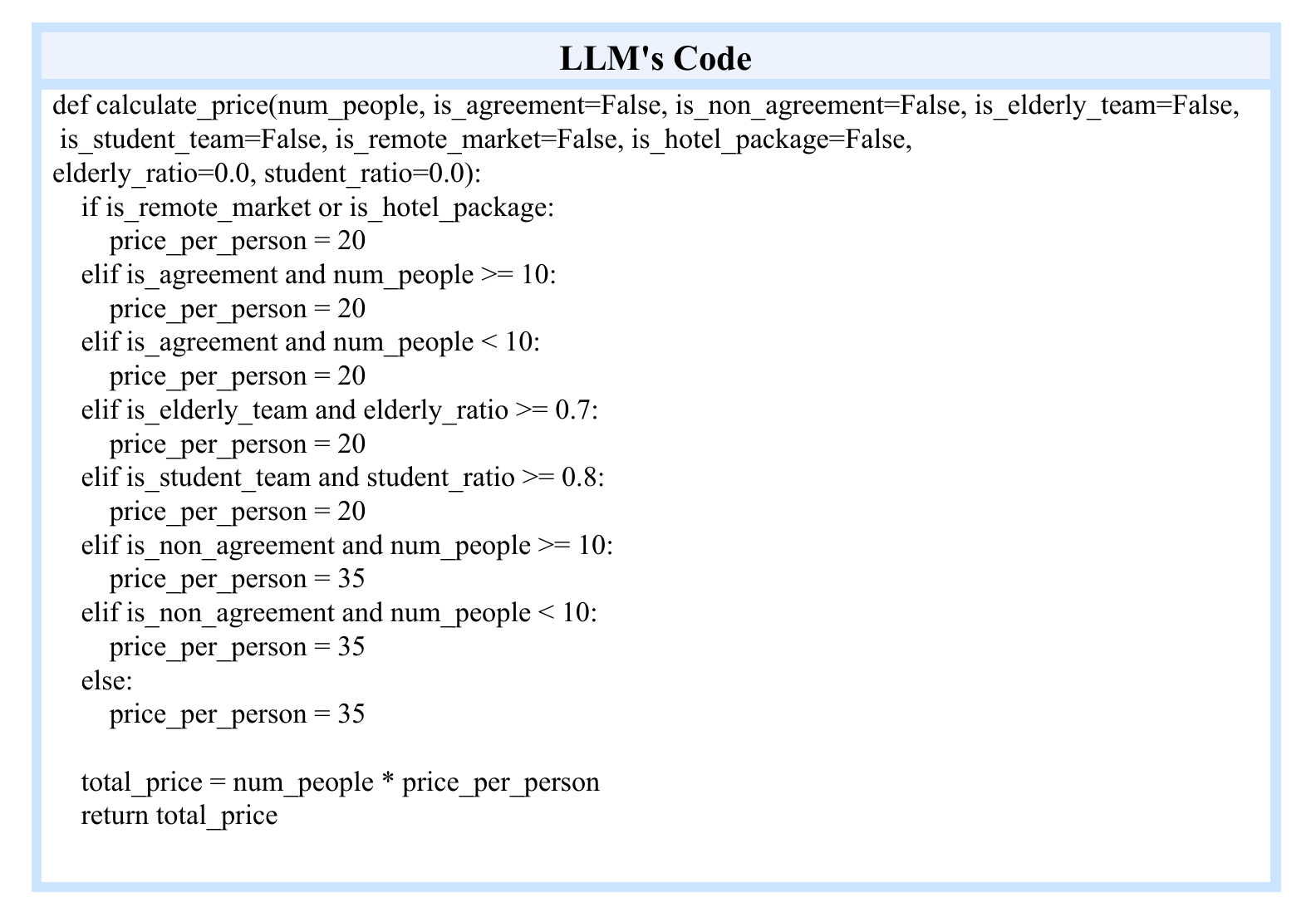}
    \caption{LLM's Code.}
    \label{fig:LLM's Code}
\end{figure}

\begin{figure}[h]
    \centering
    \includegraphics[height=0.6\textwidth]{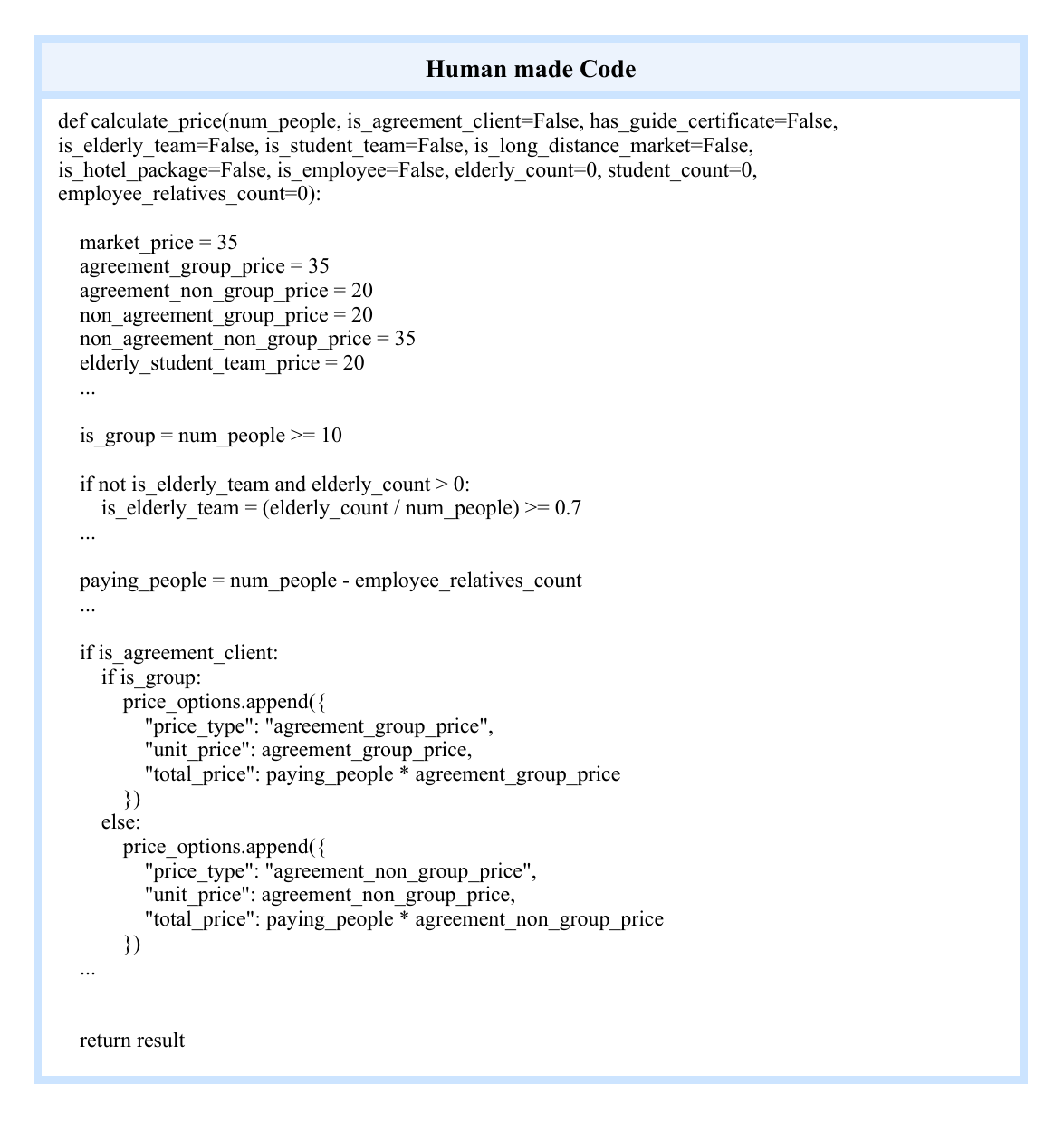}
    \caption{Human made Code.}
    \label{fig:Human made Code}
\end{figure}

The comparison between LLM-generated and human-written pricing calculation code reveals fundamental differences that directly impact the CaR method's effectiveness. \Cref{fig:LLM's Code} presents a representative LLM-generated calculator, while \Cref{fig:Human made Code} shows the corresponding human-implemented version.

The LLM-generated code follows a simplified linear if-elif structure that processes pricing rules sequentially, returning the final total cost.In contrast, the human-written code implements a sophisticated multi-option evaluation system that identifies all applicable pricing schemes, compares them systematically, and selects the optimal solution. This approach correctly handles complex conditional logic, provides detailed calculation breakdowns, and manages special cases like employee exemptions that affect the paying customer count.

\begin{table}[h]
\centering

\renewcommand\tabcolsep{5.0pt}  %
\renewcommand{\arraystretch}{1.0}
    
\begin{tabular}{lc}
\hline
\textbf{Type} & \textbf{Price} \\
\hline
Regular retail price & 80 \\
Contracted group price & 50 \\
Contracted non-group price & 50 \\
Non-contracted group price & 64 \\
Non-contracted non-group price & 72 \\
Senior/Student group price & 40 \\
Long-distance and new market price & 30 \\
Accommodation package price & 50 \\
Travel employee price & 50 \\
Free admission with hotel stay & 0 \\
\hline
\end{tabular}
\caption{Example Customer Type Price of one Attraction}
\label{tab:customer type price}
\end{table}

\begin{table}[ht]
\small
\centering
\renewcommand\tabcolsep{5.0pt}  %
\renewcommand{\arraystretch}{1.0}
\begin{tabular}{>{\raggedright\arraybackslash}p{0.35\columnwidth}>{\raggedright\arraybackslash}p{0.55\columnwidth}}
\hline
\textbf{Customer Type} & \textbf{Definition} \\
\hline
Regular retail price & Standard price for individual visitors \\
Group price & Applies when the number of visitors is $\geq$ 10 people \\
Contracted group price & Discounted rate for customers with a signed contract \\
Contracted non-group price & Price for contracted customers who don't meet group size requirement \\
Non-contracted group price & Group rate for customers without a contract (requires tour guide certificate) \\
Non-contracted non-group price & Standard price for customers without a contract \\
Senior group price & Applies when seniors (55+) constitute $>$ 70\% of the group \\
Student group price & Applies when students constitute $>$ 80\% of the group \\
Long-distance market price & Special price for visitors from outside Somerset, Hampshire, and London \\
Accommodation package price & Preferential prices for contracted groups staying at designated hotels in Clayton Castle \\
Travel employee price & Special price for travel employees and companions; applies to entire group when led by an employee \\
Free admission with hotel stay & Visitors at designated Clayton Castle hotels receive free admission to select attractions \\
\hline
\end{tabular}
\caption{Customer Type Definitions}
\label{tab:customer_types}
\end{table}

\clearpage

\end{document}